%% file: main_IV2026.tex
\documentclass[letterpaper, 10 pt, conference]{ieeeconf}  

\IEEEoverridecommandlockouts                              

\usepackage{acro}
\usepackage{graphicx}

\usepackage{amsmath}
\usepackage{amssymb}
\usepackage{bm}
\usepackage{siunitx}
\usepackage{makecell}
\usepackage{multirow}
\usepackage{multicol}
\usepackage{caption} 
\usepackage{subcaption}
\usepackage{booktabs} 
\usepackage{tikz}
\usetikzlibrary{decorations.pathreplacing}
\usetikzlibrary{shapes.geometric, arrows, positioning, calc, decorations.pathreplacing}
\usepackage{hyperref}
\usepackage{cite}

\usepackage{graphicx}   
\usepackage{tabularx}   
\usepackage{array}
\usepackage{xcolor}     
\usepackage{pgfplots}
\usepackage{pgfplotstable}
\usepgfplotslibrary{groupplots}
\usepackage{tikz, pgfplots}
\usepackage{pgfplotstable}
\usepgfplotslibrary{colorbrewer}
\usetikzlibrary{pgfplots.statistics, pgfplots.colorbrewer} 
\usepackage{float} 
\usepgfplotslibrary{fillbetween}
\usetikzlibrary{fadings}
\usetikzlibrary{fit, backgrounds, scopes,shadings}
\pgfplotsset{compat=1.18}
\usepackage{filecontents}

\usepackage[symbol]{footmisc}
\graphicspath{ {./images/} }

\pgfmathdeclarefunction{gauss}{3}{%
	\pgfmathparse{1/(#3*sqrt(2*pi))*exp(-((#1-#2)^2)/(2*#3^2))}%
}

\input{markos/color_definitions}

\title{\LARGE \bf

Uncertainty-Aware Diffusion Model for Multimodal Highway Trajectory Prediction via DDIM Sampling

}

\author{Marion Neumeier${}^{1}$, Niklas Ro\ss berg\,${}^{1}$, Michael Botsch\,${}^{1}$, Wolfgang Utschick\,${}^{2}$
\thanks{$^{1}$\,Technische Hochschule Ingolstadt, 85049 Ingolstadt, Germany
        {\tt\small firstname.lastname@thi.de}}%
\thanks{$^{2}$\,Technische Universität München, 80333 München, Germany, {\tt\small utschick@tum.de}}
}
\input{acronyms}

\usepackage{textcomp}
\newcommand\copyrighttext{%
	\footnotesize \textcopyright 2026 IEEE. Personal use of this material is permitted. Permission from IEEE must be obtained for all other uses, in any current or future media, including reprinting/republishing this material for advertising or promotional purposes, creating new collective works, for resale or redistribution to servers or lists, or reuse of any copyrighted component of this work in other works. DOI: t.b.d}
\newcommand\copyrightnotice{%
	\begin{tikzpicture}[remember picture,overlay]
	\node[anchor=south,yshift=10pt] at (current page.south) {\fbox{\parbox{\dimexpr\textwidth-\fboxsep-\fboxrule\relax}{\copyrighttext}}};
	\end{tikzpicture}%
}
\begin{document}
\bstctlcite{IEEEexample:BSTcontrol}
\maketitle
\copyrightnotice
\thispagestyle{empty}
\pagestyle{empty}

\begin{abstract}
Accurate and uncertainty-aware trajectory prediction remains a core challenge for autonomous driving, driven by complex multi-agent interactions, diverse scene contexts and the inherently stochastic nature of future motion. Diffusion-based generative models have recently shown strong potential for capturing multimodal futures, yet existing approaches such as cVMD suffer from slow sampling, limited exploitation of generative diversity and brittle scenario encodings.

This work introduces cVMDx, an enhanced diffusion-based trajectory prediction framework that improves efficiency, robustness and multimodal predictive capability. Through DDIM sampling, cVMDx achieves up to a 100$\times$ reduction in inference time, enabling practical multi-sample generation for uncertainty estimation. A fitted Gaussian Mixture Model further provides tractable multimodal predictions from the generated trajectories. In addition, a CVQ-VAE variant is evaluated for scenario encoding. Experiments on the publicly available highD dataset show that cVMDx achieves higher accuracy and significantly improved efficiency over cVMD, enabling fully stochastic, multimodal trajectory prediction.
\end{abstract}
\vspace{-0pt}

\section{Introduction}
Trajectory prediction is a core component of autonomous driving systems \cite{madjid2025rev}. Yet the task is intrinsically difficult, as future motion arises from a combination of vehicle dynamics, social interactions among agents and context-dependent constraints. Furthermore, real-world driving behavior is highly stochastic and often multimodal, exhibiting significant variability across situations. In this setting, multimodality refers to the existence of several qualitatively different yet plausible future maneuvers, such as accelerating or changing lanes, given the same scenario. These challenges have encouraged the adoption of data-driven models for trajectory prediction~\cite{Deo.2018, Messaoud2021, Neumeier2021, NEIMEIERGFFT}.

Modern trajectory prediction models increasingly aim not only to produce accurate future motion estimates but also to represent the inherent uncertainty arising from multiple plausible future behaviors \cite{Manas2025, LiGuopeng2024, Neumeier2024}. Generative models are particularly well suited to this objective, as they approximate underlying data distributions and naturally represent probabilistic, multimodal outcomes. Within this class, diffusion-based generative models \cite{Ho.2020, Ho.2022} have shown strong performance due to their ability to generate diverse and coherent trajectory samples. Building on this idea, Neumeier et al. \cite{Neumeier2024} introduce cVMD, which conditions the diffusion process on discrete scenario representations learned via a VQ-VAE \cite{Oord2017}, enabling high-level behavioral context to guide prediction.

Despite its strengths, cVMD faces several limitations. First, although diffusion models inherently support multi-sample generation, cVMD produces only a single trajectory at inference time, limiting its ability to express uncertainty and multimodality. Second, the iterative nature of diffusion sampling is computationally costly, making multi-sample prediction impractical for real-time applications. Finally, scenario representations rely on a VQ-VAE, a model that is susceptible to codebook collapse \cite{Zheng2023, roßberg2025}, potentially reducing embedding diversity and robustness.

These limitations motivate extensions that preserve the expressiveness of diffusion models while improving sampling efficiency and stabilizing scenario encoding. Building on cVMD \cite{Neumeier2024}, this work proposes cVMDx\footnote{\href{https://github.com/MB-Team-THI/conditioned-vehicle-motion-diffusion-x}{github.com/MB-Team-THI/conditioned-vehicle-motion-diffusion-x}}, an enhanced framework that enables efficient and multimodal trajectory prediction. Fig.~\ref{fig:multimodal_prediction} illustrates an example of multimodal forecasts generated by the proposed method.

\input{tikz-figures/multi_modality_plot}
The key contributions of this work include:
\begin{itemize}
\item \emph{Enhanced Scenario Representation:} Integration of a CVQ-VAE \cite{Zheng2023} to mitigate codebook collapse.
\item \emph{Fast Diffusion Inference:} Adoption of DDIM sampling \cite{Song2021}, achieving up to $100\times$ faster inference than DDPM \cite{Ho.2020} and enabling practical multi-sample prediction.
\item \emph{Explicit Multimodal Modeling:} Fitting a Gaussian Mixture Model (GMM) to generated samples for mode extraction and clear representation of multiple future hypotheses.
\item \emph{Stabilized Training Objective:} Use of a velocity-based objective \cite{Salimans2022} to improve training stability and sample consistency.
\item \emph{Benchmarking:} Evaluation of prediction performance on the publicly available highD dataset \cite{Krajewski2018}.
\end{itemize}

\section{Related Work}
Trajectory prediction in autonomous driving has been extensively studied across a range of modeling paradigms, from rule-based systems \cite{Houenou2013, Werling2010} to deep learning approaches \cite{Deo.2018, Messaoud2021, HSTA.2021, LI2023110990, Fertig2025, Neumeier.ITSC2023}. Developments in this area are thoroughly reviewed in \cite{madjid2025rev}. 

Recent work on trajectory prediction increasingly emphasizes not only accuracy, but also the modeling of multimodality and uncertainty \cite{JiangCVPR2023, HuangCUQDS2025}. In safety-critical contexts, such as autonomous driving, uncertainty-aware prediction is essential for managing ambiguous situations and enabling risk-sensitive planning. As a result, many approaches now integrate uncertainty modeling to better reflect the inherent unpredictability of agent behavior. Consequently, both probabilistic~\cite{Wiest2012, Bok2024, DESHPANDE2024104} and generative approaches~\cite{Egolf2025, YanGenRev2025, HuangRehao2020, LI2023110990} have been explored. These methods aim to generate well-calibrated and diverse trajectory distributions that reflect the stochastic nature of real-world driving environments.

Among generative approaches, diffusion models \cite{Sohl-Dickstein2015, Ho.2020, Ho.2022} have recently emerged as a powerful framework for trajectory prediction \cite{BaecVPR2024, LI2023110990, ma2025diff}. Several works have adapted diffusion models to autonomous driving by introducing task-specific conditioning strategies and architectural designs. Jiang et al.~\cite{JiangCVPR2023} proposed MotionDiffuser, which incorporates past motion, scene context and agent interactions using a permutation-invariant transformer denoiser with cross-attention. Barquero et al.~\cite{Barquero2023} introduced BeLFusion, conditioning the sampling process on a behavioral latent space to produce coherent human motion. Yuan et al.~\cite{Yuan2023} presented PhysDiff, integrating a physics-guided motion projection module and simulator to enforce physical plausibility during denoising. The conditioning signal is obtained from a discrete scenario representation learned via a VQ-VAE.

Neumeier et al.~\cite{Neumeier2024} introduced the cVMD architecture, which extends diffusion-based trajectory prediction with explicit uncertainty estimation by conditioning the model on a discrete scenario embedding derived via a VQ-VAE~\cite{Oord2017}. To ensure physically plausible predictions, the diffusion process is performed over physical control variables (acceleration and yaw rate) and trajectories are subsequently reconstructed using a Vehicle Motion Model (VMM). A key contribution of their approach is an adaptive guidance scaling mechanism that replaces the typically fixed guidance scale with an uncertainty-aware variant, strengthening conditioning in familiar scenarios and reducing it in uncertain ones. While cVMD effectively models diverse trajectories and incorporates scenario-level uncertainty, it also exhibits several limitations. 
At inference, cVMD outputs only a single trajectory, preventing the model from capturing the inherently multimodal structure of future motion and limiting its ability to express uncertainty. Moreover, the reverse-diffusion procedure introduces considerable computational overhead, making the generation of multiple behavior hypotheses impractical.
A further concern arises from the VQ-VAE~\cite{Oord2017}–based scenario module, which is susceptible to codebook collapse \cite{Zheng2023, roßberg2025}. This failure mode diminishes the diversity of learned scenario embeddings. Recent findings by Roßberg et al.~\cite{roßberg2025} indicate that the CVQ-VAE~\cite{Zheng2023} mitigates this issue, providing motivation for its integration into cVMDx to improve scenario embedding robustness and prediction reliability.

These issues highlight the need for approaches that maintain the generative strengths of diffusion models while improving scenario representation and uncertainty modeling.

\section{Preliminaries}
\label{Sec:Prelim}
Denoising Diffusion Probabilistic Models (DDPMs) are a class of generative models that operate by learning to reverse a gradual noising process~\cite{Ho.2020}.
\subsection{Forward Diffusion Process}
DDPMs assume a Markovian forward process in which data is gradually perturbed by Gaussian noise, producing a sequence $\bm{x}_0 \rightarrow \bm{x}_1 \rightarrow \cdots \rightarrow \bm{x}_T$ that converges to an isotropic Gaussian $\bm{x}_T \sim \mathcal{N}(\mathbf{0}, \mathbf{I})$. Concretely, the forward noising process is defined as the Markov chain
\begin{equation}
    q(\bm{x}_t \mid \bm{x}_{t-1}) = \mathcal{N}\!\left(\sqrt{\alpha_t}\,\bm{x}_{t-1},\, (1 - \alpha_t) \mathbf{I}\right),
\end{equation}
initialized from data samples $\bm{x}_0 \sim p_{\text{data}}$. This admits the closed-form expression
\begin{equation}
    \bm{x}_t = \sqrt{\bar{\alpha}_t}\, \bm{x}_0 + \sqrt{1 - \bar{\alpha}_t}\, \epsilon, 
    \qquad \epsilon \sim \mathcal{N}(\mathbf{0}, \mathbf{I}),
\end{equation}
where $\bar{\alpha}_t = \prod_{s=1}^t \alpha_s$ and $\alpha_t = 1 - \beta_t$.
Here, $\sqrt{\bar{\alpha}_t}$ is the cumulative signal scaling factor from the forward noising process
and $\sigma_t^2= 1 - \bar{\alpha}_t$ is the associated noise variance at step $t$ as specified by the noise schedule.
For sufficiently large $T$, the distribution of $\bm{x}_T$ approaches $\mathcal{N}(\mathbf{0}, \mathbf{I})$.

\subsection{Reverse Diffusion: Model Parameterizations}
While the forward diffusion process is analytically tractable, the reverse
conditionals $q(\bm{x}_{t-1} \mid \bm{x}_t)$ are generally intractable. Diffusion models
therefore introduce a learnable reverse process $p_{\boldsymbol{\theta}}(\bm{x}_{t-1} \mid \bm{x}_t)$
to approximate these transitions. Intuitively, $p_{\boldsymbol{\theta}}$ is trained to
denoise $\bm{x}_t$ by recovering the remaining signal structure originating from the
clean data sample $\bm{x}_0$. Once learned, this enables sampling by drawing 
$\bm{x}_T \sim \mathcal{N}(\mathbf{0}, \mathbf{I})$ and iteratively denoising back to $\bm{x}_0$. The reverse process is modeled as a Gaussian transition
\begin{equation}
p_{\boldsymbol{\theta}}(\bm{x}_{t-1}\mid \bm{x}_t)
= \mathcal{N}\!\big(
\boldsymbol{\mu}_{\boldsymbol{\theta}}(\bm{x}_t, t),
\boldsymbol{\Sigma}(t)
\big),
\end{equation}
where typically the covariance $\boldsymbol{\Sigma}(t)$ is fixed according to the noise schedule
(e.g., $\boldsymbol{\Sigma}(t) = \tilde{\beta}_t \mathbf{I}$, with $\tilde{\beta}_t = \frac{(1- \bar{\alpha}_{t-1})}{(1-\bar{\alpha}_t)} (1-\alpha_t)$ \cite{Ho.2020}), while the mean
$\boldsymbol{\mu}_{\boldsymbol{\theta}}(\bm{x}_t, t)$ is determined by a neural network
$f_{\boldsymbol{\theta}}(\bm{x}_t, t)$. Using the closed-form posterior $q(\bm{x}_{t-1}\mid \bm{x}_t,\bm{x}_0)$, the reverse update can be expressed as
\begin{equation}
\bm{x}_{t-1}
=
\underbrace{
\frac{1}{\sqrt{\alpha_t}}
\left(
\bm{x}_t
-
\frac{1-\alpha_t}{\sqrt{1-\bar{\alpha}_t}}\,
\bm{\epsilon}_{\boldsymbol{\theta}}(\bm{x}_t,t)
\right)
}_{\boldsymbol{\mu}_{\boldsymbol{\theta}}(\bm{x}_t,t)}
+\sqrt{\tilde{\beta}_t}\,\bm{z},
\end{equation}
where $\bm{z}\sim\mathcal{N}(\mathbf{0},\mathbf{I})$.
The network $f_{\boldsymbol{\theta}}(\bm{x}_t,t)$ may be trained to predict either the added noise
$\bm{\epsilon}$ (\emph{noise prediction}) or the clean signal $\bm{x}_0$ (\emph{data prediction}),
from which the reverse mean $\boldsymbol{\mu}_{\boldsymbol{\theta}}(\bm{x}_t,t)$ can be reconstructed
in closed form.

However, both noise- and data-prediction parameterizations involve scale factors
$\alpha_t$ and $\sigma_t$ that vary significantly across timesteps, which can lead to
timestep-dependent imbalance in the learning signal. To mitigate this, the
velocity parameterization \cite{Salimans2022} introduces a time-consistent target and
trains the model to predict it via
\begin{equation}
    \mathcal{L}_{\text{vel}}(\boldsymbol{\theta})
    =
    \mathbb{E}_{t,\,\bm{x}_0,\,\bm{\epsilon}}
    \big[
        \|\, \bm{v}_t - \bm{v}_{\boldsymbol{\theta}}(\bm{x}_t, t) \|_2^2
    \big],
    \label{eq:velloss}
\end{equation}
\begin{equation}
    \bm{v}_t = \sqrt{\bar{\alpha}_t}\, \bm{\epsilon} - \sigma_t\, \bm{x}_0 
\end{equation}

The variable $\bm{v}_t$ provides a well-conditioned interpolation between data and noise.
$\bm{x}_0$ and $\bm{\epsilon}$ can be recovered via
\begin{equation}
\bm{x}_0 = \sqrt{\bar{\alpha}_t}\,\bm{x}_t - \sigma_t\,\bm{v}_t,
\qquad
\bm{\epsilon} = \frac{1}{\sigma_t}\big(\bm{x}_t - \sqrt{\bar{\alpha}_t}\,\bm{x}_0\big).
    \label{eq:velrec}
\end{equation}
This parameterization improves stability and yields more uniform training across timesteps \cite{Salimans2022}.

\subsection{Classifier-free guidance (CFG)}
In diffusion models, guidance refers to controlling the generation process by incorporating additional conditioning information. Classifier-free guidance (CFG) \cite{Ho.2022} enables this without requiring an external classifier by jointly training a conditional noise estimator $\epsilon_{\bm{\theta}}(\bm{x}_t, t,  c)$ and an unconditional estimator $\epsilon_{\bm{\theta}}(\bm{x}_t, t)$ within the same network ($c=\emptyset$).
During sampling, the guided noise estimate is computed as
\begin{equation}
    \tilde{\epsilon}_{\bm{\theta}}(\bm{x}_t, c, t)
    = (1 + w)\,\epsilon_{\bm{\theta}}(\bm{x}_t, t,  c) - w\,\epsilon_{\bm{\theta}}(\bm{x}_t, t), \label{eq:combindedeps}
\end{equation}
where $w \in \mathbb{R}$ is the guidance scale. The scale $w$ balances fidelity and diversity: larger $w$ increases adherence to the conditioning (higher fidelity, lower diversity), while smaller $w$ yields more diverse but less condition-aligned samples.

\section{Methodology}
\renewcommand*{\thefootnote}{\fnsymbol{footnote}}
\renewcommand\footnoterule{\rule{0.4\linewidth}{0.8pt}}
\begin{figure*}[t!]
\vspace{2pt}
\input{tikz-figures/cvmd_pipeline}
\vspace{-2pt}
\caption[Caption in ToC]{The cVMDx architecture consists of a vehicle-motion diffusion module and a context-conditioning module. The latter, implemented as a CVQ-VAE, discretizes the traffic scenario $\boldsymbol{\xi}$, determining a codebook index $q$ used as the diffusion condition $c=q$ for the trajectory prediction. Scenario uncertainty $\delta$ from the uncertainty quantification (UQ) unit adaptively adjusts the guidance scale $w$ during generation. Red elements denote training; green denote evaluation.
}
\label{fig:architecture}
\vspace{-12pt}
\end{figure*}
This section introduces cVMDx, an enhanced diffusion-based trajectory prediction framework that refines and builds upon cVMD~\cite{Neumeier2024}. The method introduces four key enhancements: (i) replacement of the VQ-VAE encoder with \mbox{CVQ-VAE}~\cite{Zheng2023} to prevent codebook collapse in scenario representation, (ii) training the diffusion model using a velocity-based objective, (iii) DDIM-based deterministic sampling for faster inference, and (iv) a cosine-guided and uncertainty-aware CFG scheme for reliable trajectory prediction. Implementation details and code: \href{https://github.com/MB-Team-THI/conditioned-vehicle-motion-diffusion-x}{https://github.com/MB-Team-THI/conditioned-vehicle-motion-diffusion-x}

\subsection{Problem Formulation}
Let $\mathcal{D} = \{(\bm{\xi}^{(m)}, \bm{Y}^{(m)}, \mathbf{s}^{(m)})\}_{m=1}^M$ denote a dataset composed of traffic scenarios where each sample contains: (i) an observed traffic scenario $\boldsymbol{\xi}^{(m)} \in \mathbb{R}^{N \times F \times T_{\text{obs}}}$,
(ii) the future trajectory of a designated target vehicle 
$\bm{Y}^{(m)} \in \mathbb{R}^{2 \times T_{\text{pred}}}$,
and (iii) the associated maneuver class label $\mathbf{s}^{(m)} \in \mathbb{R}^{3}$, which is one-hot encoded to indicate whether the maneuver corresponds to a lane change to the left (lcl), lane change to the right (lcr), or lane keeping (kl). The observation $\boldsymbol{\xi}^{(m)}$ describes $N = 9$ interacting vehicles over a time window of $T_{\text{obs}} = 3\,\text{s}$. Each vehicle is encoded using $F = 4$ features forming
$\boldsymbol{\xi}_j^{(m)} = [x_j, y_j, v_{j,x}, v_{j,y}]^\top,$
with longitudinal and lateral positions $(x, y)$ and velocities $(v_x, v_y)$.
The prediction target is the future trajectory of the target vehicle,
\mbox{$\bm{Y}^{(m)} = [\bm{x}_{\text{pred}}, \bm{y}_{\text{pred}}]^\top$}
over a $T_{\text{pred}} = 5\,\text{s}$ horizon.

\subsection{Architecture Overview}
The proposed model consists of two core components: a \emph{context conditioning module} and a \emph{trajectory prediction module}. The context module is implemented as a CVQ-VAE, which encodes the observed motion history ($T_{\text{obs}} = \SI{3}{\second}$) of a traffic scenario into a discrete scenario token \mbox{$q \in \{1, \dots, Q\}$} from a learned codebook. This discrete scenario representation $q$ subsequently serves as the conditioning input to the diffusion-based trajectory prediction module. To enable uncertainty-aware predictions, an adaptive guidance mechanism modulates the conditioning strength throughout sampling. This modulation places stronger emphasis on contextual cues in familiar scenarios while allowing greater flexibility when scenario uncertainty is high. An overview of the full architecture is shown in Fig.~\ref{fig:architecture}.

\subsection{Context Conditioning via CVQ-VAE}
The context conditioning module discretizes the space of all observed traffic scenarios
$\boldsymbol{\xi}^{(m)} \in \mathbb{R}^{N \times F \times T_{\text{obs}}}$ into a finite set of $Q$ scenario categories.
In this work, the module is implemented using a CVQ-VAE \cite{Zheng2023}, which encodes each scenario into a latent feature
$\hat{\bm{z}}^{(m)} = E(\bm{\xi}^{(m)})$ and quantizes it to the nearest codebook entry
\vspace{-2pt}
\begin{equation}
    \bm{z}_q^{(m)} = \arg \min_{\bm{z}_k \in \mathcal{Z}} \|\hat{\bm{z}}^{(m)} - \bm{z}_k\|_2^2,
\end{equation}
where $\mathcal{Z} = \{\bm{z}_1, \dots, \bm{z}_Q\}$ is a learned codebook of scenario prototypes.
The decoder reconstructs the input \mbox{$\bm{\hat{\xi}}^{(m)}=D(\bm{z}_q^{(m)})$} and the model is trained with the standard VQ-VAE reconstruction and quantization losses \cite{Oord2017}
{
\vspace{-2pt}
\thinmuskip=1mu
\medmuskip=2mu plus 1mu minus 2mu
\thickmuskip=1mu plus 1mu
\begin{align}
\mathcal{L}_{\text{cvq}}
=
    \|\bm{\xi}^{(m)} - \bm{\hat{\xi}}^{(m)}\|_2^2
    &+ 
    \|\text{sg}[E(\bm{\xi}^{(m)})] - \bm{z}_q^{(m)}\|_2^2\notag\\
    &+
    \beta \|E(\bm{\xi}^{(m)}) - \text{sg}[\bm{z}_q^{(m)}]\|_2^2.
\end{align}
}
CVQ-VAE extends VQ-VAE by adaptively updating codebook entries to prevent codebook collapse, promoting balanced usage of scenario tokens and improved latent capacity~\cite{Zheng2023}. This effect has been observed empirically in the context of traffic scenario encoding by Roßberg et al.~\cite{roßberg2025}. 
To encourage the latent representation to encode maneuver-type information, a lightweight linear classifier is applied to $\bm{z}_q^{(m)}$ and optimized via a cross-entropy loss similar to \cite{Neumeier2024}
\begin{align}
\mathcal{L}_\textrm{cl} &= - \sum^{S}_{i=1} {s}_i^{(m)} \log({p}_i^{(m)}),
\end{align} 
where $\bm{p}^{(m)} = \textrm{softmax}(f_{cl}(\bm{z}_q^{(m)}))$. Hence, the overall objective of the module yields
\begin{equation}
    \mathcal{L}= \mathcal{L}_{\text{cvq}} + \lambda\,\mathcal{L}_{\text{cl}},
\end{equation}
where $\lambda$ is a scalar hyperparameter that controls the influence of the classification loss $\mathcal{L}_{\text{cl}}$.
After training, each codebook entry $\bm{z}_q$ corresponds to a distinct traffic scenario pattern. Consequently, the large and heterogeneous space of scenario constellations is discretized into $Q$ clusters. Each cluster $q$ groups together scenarios that exhibit comparable contextual attributes, providing a discrete and interpretable basis for conditioning the diffusion model.

\subsubsection*{Scenario Context Uncertainty Estimation}
The traffic scenario context uncertainty is determined in the discrete latent space of the CVQ-VAE, similar to the approach in \cite{Neumeier2024}. This provides an estimate of how well a given scenario aligns with the characteristics of its assigned scenario cluster $q$. Therefore, after training, each latent embedding $\hat{\bm{z}}^{(m)}$ is assigned to its nearest codebook entry $\bm{z}_q$, forming clusters \mbox{$\mathcal{H}_q = \{ {\hat{\bm{z}}}^{(1)}, \hat{\bm{z}}^{(2)},\dots, \hat{\bm{z}}^{(h_q)} \}$} of similar scenarios. For each cluster $q$, a multivariate Gaussian \mbox{$\mathcal{N}(\bm{\mu}_q, \bm{\Sigma}_q)$}, is fitted by Maximum Likelihood, yielding
{
\thinmuskip=1mu
\medmuskip=2mu plus 1mu minus 2mu
\thickmuskip=1mu plus 1mu
\begin{equation}
\bm{\mu}_q = \bm{z}_q,\quad 
\bm{\Sigma}_q = \mathbb{E}_{\hat{\bm{z}}^{(h)} \in \mathcal{H}_q}\!\left[(\hat{\bm{z}}^{(h)} - \bm{\mu}_q)(\hat{\bm{z}}^{(h)} - \bm{\mu}_q)^\top\right].
\end{equation}}
For any sample $\hat{\bm{z}}^{(m)}$ assigned to cluster $q$, its alignment with the cluster is quantified via the Mahalanobis distance
\begin{equation}
\delta_m = \sqrt{(\hat{\bm{z}}^{(m)} - \bm{\mu}_q)^\top \bm{\Sigma}_q^{-1} (\hat{\bm{z}}^{(m)} - \bm{\mu}_q)},
\end{equation}
a covariance-aware metric.
A larger $\delta_m$ indicates that the sample lies farther from the typical cluster region and is therefore treated as more uncertain with respect to the scenario context. This estimate is used to modulate the conditioning strength during trajectory generation: By adapting the guidance scale $w$ depending on $\delta_m$, this contextual uncertainty is directly integrated into the diffusion dynamics, influencing how strongly the model adheres to the scenario conditioning during trajectory generation. The specific adaptation scheme for $w$ is detailed in the subsequent section.

\subsection{Vehicle Motion Diffusion (VMD): Training}
This module performs trajectory prediction using a generative classifier-free guided DDPM.
Its training is decoupled from the context conditioning module and relies on the tokens $q \in \{1, \dots, Q\}$ produced by the pretrained CVQ-VAE. In contrast to the context conditioning
module, the VMD module does not receive the full observed scenario context~$\bm{\xi}^{(m)}$.
Instead, it is conditioned only on the discrete scenario index $c^{(m)} = q$, which serves
as the conditioning signal to the DDPM. Because $c^{(m)}$ is a scenario category index rather
than a single deterministic scenario representation, the model learns a distribution of feasible motion patterns instead of a single most likely trajectory. This
captures the intrinsic variability in driver behavior, where similar traffic situations may
lead to different but plausible future motions.

Rather than predicting future waypoints directly, the DDPM generates a sequence of vehicle
control inputs $\hat{\bm{x}}^{(m)}_{0} = [\bm{\hat{a}}^{(m)}, \hat{\dot{\bm{\psi}}}_{}^{(m)}]$, consisting of longitudinal
acceleration and yaw-rate commands. These control inputs are used for training the model. 
During the inference phase, however, these controls are passed through a VMM to obtain the predicted trajectory
$\hat{\bm{Y}}^{(m)} = [\hat{\bm{x}}_{\text{pred}}, \hat{\bm{y}}_{\text{pred}}]^\top\, \bm{\hat{y}} \in \mathbb{R}^{T_{\text{pred}}}$, ensuring physical
plausibility by construction.

\subsubsection*{Training via Velocity Objective}
Let $\bm{x}_0^{(m)} \in \mathbb{R}^{2\times T_\textrm{pred}}$ denote the ground-truth longitudinal acceleration and yaw-rate control sequence. The training objective is the
mean-squared velocity reconstruction loss Eq.~\ref{eq:velloss} (cf. Sec. \ref{Sec:Prelim}). During training, the model is conditioned on the corresponding scenario index $c^{(m)} = q$. The reverse diffusion process is parameterized by a U-Net~\cite{Ronneberger2015}, which
learns to approximate the denoising transitions $p_{\boldsymbol{\theta}}(\bm{x}_{t-1}\mid\bm{x}_t, c^{(m)})$ and thereby generate dynamically consistent and context-adaptive motion controls. To enable CFG, the conditioning input is randomly omitted during training with a fixed probability of 10\%. This trains the network to perform both conditional and unconditional denoising, enabling controllable generation during inference by combining the two predictions~\cite{Ho.2022}.

\subsection{Vehicle Motion Diffusion (VMD): Inference}
\subsection*{Uncertainty-Aware Classifier-Free Guidance}
During inference, the model predicts both a conditional $\bm{v}_{\boldsymbol{\theta}}(\bm{x}_t, t, c)$ and an unconditional denoising output $\bm{v}_{\boldsymbol{\theta}}(\bm{x}_t, t, \varnothing)$. These outputs are first mapped from velocity space to the noise domain via
\begin{equation}
\hat{\epsilon}_{\bm{\theta}}(\bm{x}_t, t,c) = \sigma_t \bm{x}_t + \alpha_t \, v_{\boldsymbol{\theta}}(\bm{x}_t, t, c), 
\end{equation}
and subsequently combined as in Eq.~\ref{eq:combindedeps}.

In standard diffusion settings, $w_t$ is held constant across sampling steps. However, a
fixed guidance scale can amplify details and reduce realism \cite{Wang2024}. 
To prevent such over-conditioning, in this work the guidance scale $w_t(\delta_m, t)$ is adapted based on both the latent-space–based estimate of scenario uncertainty and the diffusion timestep.
A cosine-based \cite{Wang2024} CFG schedule controls the guidance scale
{
\thinmuskip=1mu
\medmuskip=2mu plus 1mu minus 2mu
\thickmuskip=1mu plus 1mu
\begin{equation}
w(t,\delta_m) = w_{\min}
+ \left(w_{\max}(\delta_m)-w_{\min}\right)
 \frac{1 - \cos\left(\frac{\pi t}{T}\right)}{2},
\end{equation}}
where $t$ denotes the diffusion timestep and
{
\thinmuskip=1mu
\medmuskip=2mu plus 1mu minus 2mu
\thickmuskip=1mu plus 1mu
\begin{equation}
w_{\max}(\delta_m) = w_{\text{max,base}}  \left(1 - \frac{\min(\delta_m, t_c)}{t_c}\right).
\end{equation}}
The threshold $t_c$ defines the point beyond which the maximal conditioning weight saturates.
The conditioning weight is adaptively modulated by the scenario distance $\delta_m$ estimated in
the CVQ-VAE latent space, which serves as an uncertainty indicator. Lower uncertainty yields stronger maximal conditioning, while higher uncertainty relaxes it to preserve diverse motion hypotheses.
A cosine schedule over the diffusion timestep $t$ prevents over-conditioning in later stages, yielding more stable and realistic samples \cite{Wang2024}. This cosine-guided uncertainty-adaptive weighting enforces stronger conditioning in familiar contexts and softer, more exploratory behavior under uncertainty, maintaining both multimodality and behavioral plausibility.

\subsection*{DDIM-based Deterministic Sampling}
A key limitation of DDPMs is sampling inefficiency: generating a sample typically requires hundreds to thousands of sequential denoising steps, resulting in high computational cost. This limitation is particularly impractical in the automotive domain, where real-time or near-real-time performance is critical, making such high computational demands infeasible.

To accelerate sampling, this work uses DDIM \cite{Song2021}, which enables faster generation with fewer diffusion steps. DDIM interprets the diffusion process as the solution of an ordinary differential equation (ODE) rather than a stochastic process, yielding a deterministic and more efficient sampling scheme\cite{Song2021}. The update from $t$ to $t-1$ is given by
\begin{equation}
\bm{x}_{t-1} = \sqrt{\alpha_{t-1}} \hat{\bm{x}}_0 + \sqrt{1 - \alpha_{t-1} - \sigma_t^2}  \epsilon_{\bm{\theta}}(\bm{x}_t, t) + \sigma_t  \eta,
\end{equation}
where $\bm{x}_0$ and $\bm{\epsilon}$ are reconstructed from the model prediction (cf. Eq.\ref{eq:velrec}). Due to its deterministic formulation, the DDIM sampling trajectory can be discretized with fewer timesteps by selecting a reduced subset from the full DDPM schedule
\begin{equation}
t_1, t_2, \dots, t_T \;\;  \Rightarrow\; \;  t'_1, t'_2, \dots, t'_S \quad \text{with } S \ll T.
\end{equation}
\subsection*{Vehicle Motion Model (VMM)}
VMMs are mathematical representations of vehicle motion kinematics. They describe the
relationship between control inputs and the resulting motion while respecting non-holonomic constraints. In this work, the motion control parameters are defined as
$
\bm{x}_0^{(m)} = [\,\bm{a}_{x}, \dot{\bm{\psi}}\,]$, where \mbox{$\dot{\bm{\psi}}, \bm{a}_x \in \mathbb{R}^{T_{\text{pred}}}$}
with $\bm{a}_{x,t}$ and $\dot{\bm{\psi}_t}$ representing the longitudinal acceleration and yaw rate at timestep $t$. Following~\cite{Neumeier2024}, the
position $(x_t, y_t)$, velocity $v_t$, and heading $\psi_t$ of a vehicle at time step
$t + \tau$ are computed as
{\thinmuskip=1mu
\medmuskip=2mu plus 1mu minus 2mu
\thickmuskip=1mu plus 1mu
\begin{align}
x_{t+\tau}\!&=\!x_{t}\!+\!v_{t}c(\psi_t)\tau\!+\!( a_{x,t} c(\psi_t)\!-\!\dot{\psi}_t v_t s(\psi_t)){\tau^2}/{2}, \label{eq:VMM1}\\	
y_{t+\tau}\!&=\!y_{t}\!+\!v_{t}s(\psi_t)v\!+\!(a_{x,t}s(\psi_t)\!+\!\dot{\psi}_t v_t c(\psi_t)){\tau^2}/{2},\\
v_{t+\tau}\!&=\!v_{t}\!+\!a_{x,t},\tau\\
\psi_{t+\tau}\!&=\!\psi_{t}\!+\!\dot{\psi}_t \tau, \label{eq:VMM2}
\end{align}}
where $c(\psi_t)=\cos(\psi_t)$, $s(\psi_t)=\sin(\psi_t)$, and $\tau$ denotes the time
increment. Based on the predicted control parameters $\bm{\hat{x}}_0^{(m)}$ for a given data sample $m$, the prediction is transformed by Eq.~\ref{eq:VMM1}--~\ref{eq:VMM2} into the trajectory prediction $\hat{\bm{Y}}^{(m)}$.

\subsection*{Multimodality Requirements}
Due to the stochastic nature of diffusion-based generative models, each sample yields a distinct trajectory realization. Relying on a single generated trajectory, as in \cite{Neumeier2024}, provides a poor estimator of the underlying predictive distribution and is insufficient for robust inference. In this work, multiple trajectories are generated for each scenario to account for prediction stochasticity and multimodality. These samples can then be aggregated by (i) averaging to obtain an expected mean trajectory or (ii) analyzed as distinct hypotheses representing alternative plausible motion outcomes.

To determine how many trajectories must be synthesized to obtain a stable mean trajectory
estimate, the confidence-interval–based sample size estimator from \cite{Hazra2017} is used:
\begin{equation}
    N_t \geq \left(\frac{z \cdot \sigma_t}{\varepsilon}\right)^2,
\end{equation}
where $z$ is the standard normal quantile associated with the desired confidence level
(e.g., $z=1.96$ for 95\% confidence), $\sigma_t$ is the empirical standard deviation of the
generated trajectories at prediction timestep $t$, and $\varepsilon$ denotes the tolerated mean estimation error.

The tolerated spatial error $\varepsilon$ is defined relative to the scale of the predicted motion. For highway trajectories of approximately $180\,\mathrm{m}$ over a $5\,\mathrm{s}$ horizon, an accuracy target of $1\text{--}2\%$ (i.e., $2.5\text{--}3\,\mathrm{m}$) is assumed to be appropriate. To estimate the variability $\sigma_t$, a pilot run of $500$ generated trajectories is used and $\sigma_t$ is approximated via
\begin{equation}
    \sigma_t = \sqrt{\mathrm{Var}(x_t) + \mathrm{Var}(y_t)}.
\end{equation}

Hence, at $t = 5\,\mathrm{s}$, with $\sigma_{5s} = 4.06\,\mathrm{m}$ and
$\epsilon = 2.81\,\mathrm{m}$:
\begin{equation}
    N_{5s} \approx \left(\frac{1.96 \cdot 4.06}{2.81}\right)^2 \approx 8.7.
\end{equation}

Thus, sampling $N_{\mathrm{samples}} = 9$ trajectories per scenario provides a sufficiently accurate estimate of the mean of the diffusion model’s predictive distribution, while keeping computational effort manageable. Accordingly, the set of trajectory predictions generated for a data sample $m$ is
$
    \widehat{\mathcal{Y}}^{(m)} 
    = \left\{ \boldsymbol{\hat{Y}}^{(i)} \right\}_{i=1}^{N_{\mathrm{samples}}}.
$

Importantly, this computation of $N_{\mathrm{samples}}$ does not measure the prediction error with respect to the ground-truth future trajectory. Instead, it determines how many samples are required so that the model’s estimated mean prediction is bounded by the desired confidence interval.

\subsection*{(i) Single Mean Trajectory:}
The mean prediction is simply the sample average
\begin{equation}
\bar{\boldsymbol{Y}}^{(m)} 
= \frac{1}{N_{\mathrm{samples}}}
  \sum_{i=1}^{N_{\mathrm{samples}}} \boldsymbol{\hat{Y}}^{(i)} .
\end{equation}

\subsection*{(ii) Motion Hypotheses Generation:}
To identify and characterize multimodal behavior from a set of sampled trajectories, a two-stage clustering pipeline is applied. First, the $N_\textrm{samples}=9$ trajectory samples are projected into a lower-dimensional space using Principal Component Analysis (PCA)
{\thinmuskip=1mu
\medmuskip=2mu plus 1mu minus 2mu
\thickmuskip=1mu plus 1mu
\begin{equation}
\hat{\bm{Y}}^{(m, i)}\in\mathbb{R}^{2 \times T_{\mathrm{pred}}}
\hspace{4pt} \mapsto \quad
\bm{z}_{\textrm{pca}}^{(m, i)}=\mathrm{PCA}\!\left(\hat{\bm{Y}}^{(m, i)}\right)
\in \mathbb{R}^{r},    
\end{equation}
}
of dimension $r=2$. This dimensionality reduction step preserves the relevant spatio-temporal variation among trajectories. This reduces noise and enables more clustering robustness.
Following PCA, a GMM \cite{Duda1974PatternCA}
\begin{equation}
    p(\bm{z}_{\textrm{pca}}) = \sum^C_{c=1} \pi_c \mathcal{N}(\bm{z}_{\textrm{pca}}|\bm{\mu}_c, \bm{\Sigma}_c) \quad \text{where} \sum_{c=1}^C \pi_c = 1,
\end{equation}
is fitted to the reduced representations to identify distinct motion hypotheses. Since the optimal number of clusters $C$ is not known a priori, multiple GMMs with varying numbers $C = {1, \dots, C_{\textrm{max}}}$ of components are trained. Model selection is then performed using the Bayesian Information Criterion (BIC) \cite{schwarz1978bic}, which evaluates each candidate model by balancing data likelihood against model complexity via
\begin{equation}
\text{BIC} = -2 \ln(L) + p \ln(N_\textrm{samples}), 
\end{equation}
where $L$ is the maximized value of the likelihood function and $p$ is the total number of learnable parameters. For a GMM with $C$ components and $r=2$, the parameter count is $p = 6C-1$.
The GMM that yields the lowest BIC score is selected as the most appropriate representation of the multimodal distribution. This selection simultaneously determines the optimal number of hypotheses $C$
(i.e., mixture components) and provides the corresponding assignments of realizations within $\widehat{\mathcal{Y}}$ to its most likely hypothesis \mbox{$\mathcal{C} \in \{1, \dots, C\}$},  $C\leq C_\textrm{max}$. Hence, each hypothesis $c$ is associated with a subset $\widehat{\mathcal{Y}}^{(m,\mathcal{C} )} \subset \widehat{\mathcal{Y}}^{(m)}$.
The expected trajectory of hypothesis $c$ is then defined as the mean
\begin{equation}
    \bar{\boldsymbol{Y}}^{(m, \mathcal{C} )} 
    = \frac{1}{|\widehat{\mathcal{Y}}^{(m,\mathcal{C} )}|}
      \sum_{\boldsymbol{\hat{Y}}^{(i)} \in \widehat{\mathcal{Y}}^{(m,\mathcal{C} )}}
      \hat{\boldsymbol{Y}}^{(i)} .
\end{equation}

This approach enables the unsupervised discovery of high-level behavioral hypotheses (e.g., braking, lane changes) directly from the sample distribution, without manual labeling or assumptions about the number of possible hypotheses. Moreover, the number of predicted trajectories assigned to each hypothesis provides an empirical measure of its relative likelihood $
p(c) \approx 
\frac{|\widehat{\mathcal{Y}}^{(m,c)}|}
     {N_{\mathrm{samples}}}
$: hypotheses supported
by more samples correspond to motion modes with higher probability.

\section{Experiments and Results}
\begin{table*}[t]
\centering
\vspace{5pt}
\input{tables/Ablation_study}
\caption{Ablation study on cVMDx trajectory prediction across different quantization configurations of the context-conditioning module, evaluated under three prediction modes: mean trajectory, most-likely hypothesis and best-$k$ selection.}
\label{tab:quantization_comparison}
\vspace{-14pt}
\end{table*}

\subsection{Dataset}
The study uses the publicly available highD dataset \cite{Krajewski2018}, comprising drone recordings of German highways at \SI{25}{\Hz}. Preprocessing follows \cite{Neumeier2024}.
Each scenario $\bm{\xi}^{(m)}$ contains up to $N = 9$ vehicles: the target vehicle and the up to eight surrounding vehicles.
For each vehicle $F = 4$ features are recorded over $T_{\text{obs}} = 75$ time steps \mbox{(\SI{3}{s})}. 
The vehicle features include the longitudinal ($\textbf{x}$) and lateral ($\textbf{y}$) positions as well as the corresponding velocities $\bm{v}_\text{x}$ and $\bm{v}_\text{y}$. 
Each scenario was categorized into one of \mbox{$S =3$} classes: \ac{kl}, \ac{lcl}, or \ac{lcr}, based on the target vehicle’s subsequent maneuver. The dataset is balanced across classes. 
The training dataset $\mathcal{D}_{\text{train}}$ contains a total of 9,841 samples, while the test dataset $\mathcal{D}_{\text{test}}$ includes 4,217 samples.

\subsection{Implementation Details}
The training of the vehicle motion diffusion module and the context conditioning module is conducted separately. Initially, the context conditioning module (CVQ-VAE) is trained using batch size 
\mbox{$B_{1}=64$}, learning rate \mbox{$lr_1=$\SI{4.5e-6}{}}  and weighting factor \mbox{$\lambda=1$} for a total number of epochs \mbox{$E_{1}=1200$}. After the CVQ-VAE training is complete, its parameters are frozen. Subsequently, the vehicle motion diffusion module (DDPM) is trained with a batch size of \mbox{$B_{2}=256$}, learning rate \mbox{$lr_{2}=$\SI{2.0e-4}{}}, and trained for \mbox{$E_{2}=2000$} epochs.
Before applying PCA and GMM clustering, the lateral and longitudinal coordinates are normalized via Min-Max-Scaling to ensure that the higher variance in the longitudinal motion direction does not dominate the representation. The maximum number of GMM components is limited to $C_{\textrm{max}}=3$.
For the DDIM sampling $S=10$. Further $t_c =50$, $w_\textrm{max, base}=1$ and $w_\textrm{min}=0.1$.

\subsection{Evaluation Metrics}
Besides the commonly used Average Displacement Error (ADE) and Final Displacement Error (FDE) \cite{madjid2025rev, Neumeier2024}, this work evaluates multimodal prediction performance via
\begin{align}
\textrm{MinADE}_k &= \frac{1}{N} \min_k \sum_n^N  \sum^T_t || \bar{\bm{Y}}^{(\mathcal{C}=k)}_{n,t} - \bm{Y}_{n,t} ||^2_2, \\
\textrm{MinFDE}_k &= \frac{1}{N} \min_k \sum_n^N  || \bar{\bm{Y}}^{(\mathcal{C}=k)}_{n,T} - \bm{Y}_{n,T}||^2_2.
\end{align}
These metrics compute the Euclidean distance between the ground truth trajectory and the closest prediction $k$ among $K=C_{\textrm{max}}$ possible predicted trajectories (hypothesis).
For DDIM sampling with $S = 10$, and $T = 1000$ in DDPM, roughly a $100\times$ speed-up is achieved w\,.r\,.t. \cite{Neumeier2024}. Since the DDIM sampling selects $S$ time steps out of the original $T$ in a linearly spaced manner, the sampling complexity becomes ${O}(S)$ instead of ${O}(T)$, effectively reducing the reverse process from $T$ denoising steps to only $S$ evaluations.

\subsection{Ablation Study: Number of Codebook Entries $Q$}
To evaluate the influence of the total number of codebook entries $Q$ in the CVQ-VAE on trajectory prediction performance, an ablation study is conducted by varying $Q$ within the model architecture. Specifically, the models are trained with $Q \in \{30,60,90,128, 256\}$, while keeping all other hyperparameters fixed. The objective of this analysis is to understand how the capacity of the latent space affects predictive accuracy of future trajectories.
The results, summarized in Table~\ref{tab:quantization_comparison}, show only marginal improvements as $Q$ increases. This suggests that simply enlarging the codebook does not necessarily produce a more informative or representative clustering of traffic-scenario types.
To  better understand this behavior, Figure~\ref{fig:kld} examines how codebook usage relates to latent-space regularization by plotting the Kullback--Leibler divergence $\textrm{KL}(p_q(\bm{Y}) \| p_{\bm{\theta},q}(\bm{Y}))$ against $\log(N_q)$ for $Q = 60$ and $Q = 256$. \( p_q(\bm{Y}) = p_{\text{data}}(\bm{Y} \mid q) \) is the subset distribution associated with cluster \(q\) and \( p_{\theta,q}(\bm{Y}) \) denotes the model distribution over predicted trajectories \( \hat{\bm{Y}} \).

Both configurations show a clear inverse relationship: frequently determined codebook entries $q$ ($N_q \uparrow$) exhibit lower divergence, indicating closer alignment between their posteriors and the model prior. The empirical fit $x^{-0.6} + 0.05$ illustrates this power-law decay. The similarity of the curves for $Q = 60$ and $Q = 256$ suggests that enlarging the codebook provides limited benefit. Since the dataset size is fixed, increasing $Q$ spreads data assignments over more entries, decreasing $N_q$ for many of them. Entries with low usage maintain higher KL divergence and therefore remain poorly aligned with the prior. Consequently, sparsely used entries provide little regularization, yielding negligible gains when increasing~$Q$.
\input{tikz-figures/dist_approx.tex}
\begin{table}[t!]
	\centering
        \include{tables/Performance_evaluation}

        \vspace{-2pt}
        \caption{Prediction performance of different state-of-the-art architectures based on the metrics ADE and FDE.}
	\label{tab:comparison}
	\vspace{-12pt}
\end{table}
\subsection{Benchmarking Against SOTA Point-Estimator Models}
The proposed method is evaluated against established trajectory prediction baselines that operate as point estimators, producing a single deterministic future trajectory per input scenario. These approaches typically optimize for average displacement error and therefore benefit from direct regression toward the mean motion pattern, which can yield lower error metrics in settings with limited multimodality. In contrast, the diffusion-based model generates a distribution of plausible future trajectories and is designed to represent uncertainty rather than collapse it. As a result, while the method does not outperform the strongest point-estimation baseline in terms of mean trajectory error, it provides multimodal predictions that better capture the range of feasible future behaviors. This distinction is particularly relevant in safety-critical planning, where expressing uncertainty is as important as minimizing average error.
Averaging multiple samples (cVMDx$^{\textrm{(mean)}}$) or selecting the most likely trajectory (cVMDx$^{p(c)}$) both outperform the single prediction of \cite{Neumeier2024} in ADE, indicating that aggregation yields more reliable future-motion estimates. No improvement is observed in FDE.

\section{Conclusions}
The proposed cVMDx framework substantially accelerates generation, achieving up to a 100$\times$ speedup over the cVMD architecture \cite{Neumeier2024} while enabling fully stochastic and multimodal prediction. The velocity-based training objective enhances optimization stability and the adaptive guidance scale helps prevent conditioning overemphasis during inference.

In contrast, the CVQ-VAE component provides only marginal gains over the standard VQ-VAE and variations in the quantization level $Q$ have limited influence on prediction quality. This suggests that deeper quantization is unlikely to be beneficial without substantially larger datasets or more semantically structured clustering methods. 

Overall, while CVQ-VAE offers little advantage in this setting, the other architectural and training improvements collectively yield notable gains in prediction accuracy and efficiency in generative modeling.

\bibliographystyle{IEEEtran}
\bibliography{Literature}

\end{document}

%% file: markos/color_definitions.tex
\definecolor{blue0}{HTML}{0077ff}
\definecolor{green0}{HTML}{00e89d}
\definecolor{orange0}{HTML}{ffb067}
\definecolor{beige0}{HTML}{dac7b5}

\definecolor{red1}{HTML}{ff8065}
\definecolor{green1}{HTML}{078344}
\definecolor{blue2}{HTML}{8eb8e7}

\definecolor{blue1}{HTML}{b2bde7}

\definecolor{green3}{HTML}{c3dcb5}
\definecolor{green2}{HTML}{6caf2d}

\definecolor{yellow1}{HTML}{ffdb93}

\definecolor{orange1}{HTML}{fdb31f}
\definecolor{brown0}{HTML}{b97100}

%% file: acronyms.tex
\DeclareAcronym{ADS}{
short = ADS ,
long = Automated Driving Systems ,
short-plural = s ,
long-plural = s 
}

\DeclareAcronym{CCP}{
short = CCP ,
long = Coupon Collector's Problem ,
short-plural = s ,
long-plural = s 
}

\DeclareAcronym{VQ}{
short = VQ,
long = Vector Quantization,
short-plural = s ,
long-plural = s 
}

\DeclareAcronym{CVQ-VAE}{
short = CVQ-VAE ,
long = Clustering Vector Quantized - Variational Autoencoder ,
short-plural = s ,
long-plural = s 
}

\DeclareAcronym{kl}{
short = kl ,
long = keep lane ,
short-plural = s ,
long-plural = s 
}

\DeclareAcronym{lcl}{
short = lcl ,
long = lane change left ,
short-plural = s ,
long-plural = s 
}

\DeclareAcronym{lcr}{
short = lcr ,
long = lane change right ,
short-plural = s ,
long-plural = s 
}

\DeclareAcronym{ML}{
short = ML ,
long = Machine Learning ,
short-plural = s ,
long-plural = s 
}

\DeclareAcronym{ODD}{
short = ODD ,
long = Operational Design Domain ,
short-plural = s ,
long-plural = s 
}

\DeclareAcronym{OOD}{
short = OOD ,
long = Out-of-Distribution ,
short-plural = s ,
long-plural = s 
}

\DeclareAcronym{PDF}{
short = PDF ,
long = probability density function ,
short-plural = s ,
long-plural = s 
}

\DeclareAcronym{VAE}{
short = VAE ,
long = Variational Autoencoder ,
short-plural = s ,
long-plural = s 
}

\DeclareAcronym{VQ-VAE}{
short = VQ-VAE ,
long = Vector Quantized - Variational Autoencoder ,
short-plural = s ,
long-plural = s 
}

%% file: tikz-figures/multi_modality_plot.tex
\begin{figure}[t]
\vspace{16pt}
\centering
\begin{tikzpicture}
\def \hvlen {6.80};
\def \hvlenmin {14.80};
\def \hvlenmax {1.80};
\def \hvwidth {0.85};
\def \lanewidth {3.4/2};

\pgfplotstableread[col sep=comma]{data/qidx_26_trajectories.csv}\datatable

\pgfplotstablecreatecol[
    create col/expr={
        (
            \thisrow{all_y_22} +
            \thisrow{all_y_103} +
            \thisrow{all_y_111} +
            \thisrow{all_y_197} +
            \thisrow{all_y_256} +
            \thisrow{all_y_453}
        ) / 5
    }
]{mean_y1}{\datatable}

\pgfplotstablecreatecol[
    create col/expr={
        (
            \thisrow{all_x_22} +
            \thisrow{all_x_103} +
            \thisrow{all_x_111} +
            \thisrow{all_x_197} +
            \thisrow{all_x_256} +
            \thisrow{all_x_453}
        ) / 6
    }
]{mean_x1}{\datatable}

\pgfplotstablecreatecol[
    create col/expr={
        (
            \thisrow{all_y_83} +
            \thisrow{all_y_197} +
            \thisrow{all_y_339} +
            \thisrow{all_y_368} +
        ) / 4
    }
]{mean_y2}{\datatable}

\pgfplotstablecreatecol[
    create col/expr={
        (
            \thisrow{all_x_83} +
            \thisrow{all_x_197} +
            \thisrow{all_x_339} +
            \thisrow{all_x_368} +
        ) / 4
    }
]{mean_x2}{\datatable}

\begin{axis}[
table/col sep=comma,
xlabel={\footnotesize $x [m]$},
ylabel={\footnotesize $y [m]$ },
width=\columnwidth,
ticklabel style={font=\footnotesize},   
height=3.5cm,
xtick={0,20,40,60,80,100, 120, 140, 160},
xticklabels={0,20,40,60,80,100,120, 140, 160},
table/col sep=comma,
ymax = 3.2,
ymin = -3.2,
xmin = -\hvlenmin-2,
xmax = 151,
legend pos=south east,
]
\addlegendimage{no markers, line width=0.5mm, red, dashed}
\addlegendentry{\scriptsize Ground Truth Trajectory}
\addplot[gray, opacity = 0.5,samples=100, domain=-25:180, dashed, line width=0.55mm, dash pattern={on 18pt off 28pt}] {\lanewidth}; 
\addplot[gray, opacity = 0.5,samples=100, domain=-25:180, dashed, line width=0.55mm, dash pattern={on 18pt off 28pt}] {-\lanewidth}; 
\addplot[gray, opacity = 0.5,samples=100, domain=-25:180, dashed, line width=0.55mm, dash pattern={on 18pt off 28pt}] {\lanewidth+2*\lanewidth}; 
\addplot[gray, opacity = 0.5,samples=100, domain=-25:180, dashed, line width=0.55mm, dash pattern={on 18pt off 28pt}] {\lanewidth+4*\lanewidth};


\foreach \i in {22, 103, 111, 197, 256, 453}{%
    \addplot [
        green1,
        opacity = 0.5,
        line width=0.6mm
    ]
    table [
        x={all_x_\i},
        y={all_y_\i},
        col sep=comma
    ] {data/qidx_26_trajectories.csv}; \label{realization1_0}
} 
\addplot[green1, line width=0.75mm]
    table[x=mean_x1, y=mean_y1] {\datatable};\label{mean1}

\foreach \i in {83, 197, 339, 368}{%
    \addplot [
        blue2,
        line width=0.6mm
    ]
    table [
        x={all_x_\i},
        y={all_y_\i},
        col sep=comma
    ] {data/qidx_26_trajectories.csv}; \label{realization2_83}
}
\addplot[blue, line width=0.75mm]
    table[x=mean_x2, y=mean_y2] {\datatable}; \label{mean2}

\addplot [line width=0.5mm, draw=red, dashed] table [x=x_pred_0, y=y_pred_0]{data/qidx_26_trajectories.csv}; \label{gt}
\addplot[] graphics[xmin=-\hvlenmin, ymin=-\hvwidth, xmax=\hvlenmax, ymax=\hvwidth]{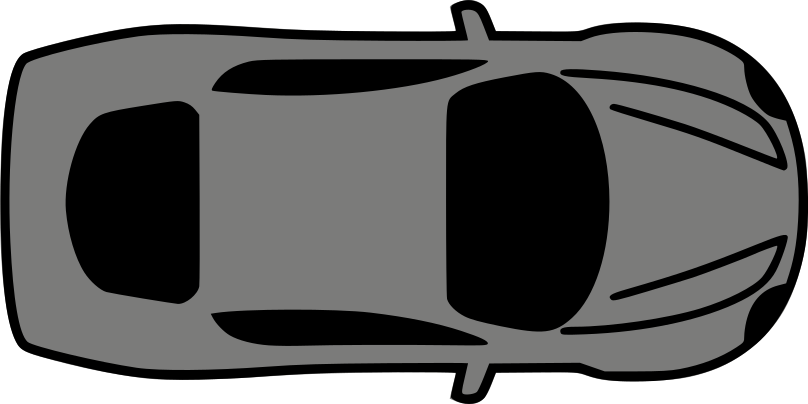};
\end{axis}
\end{tikzpicture}
\vspace{-4pt}
\caption[Caption in ToC]{
Example of multimodal trajectory prediction from cVMDx based on $N_{\mathrm{samples}}=9$ sampled trajectories. 
Modality~1 is shown in green (\ref{realization1_0}) with mean 
$\bar{\boldsymbol{Y}}^{(\mathcal{C}=1)}$ (\ref{mean1}), and modality~2 in blue 
(\ref{realization2_83}) with mean $\bar{\boldsymbol{Y}}^{(\mathcal{C}=2)}$ (\ref{mean2}).}
\label{fig:multimodal_prediction}
\vspace{-13pt}
\end{figure}

%% file: tikz-figures/cvmd_pipeline.tex
\pgfplotsset{
	matrix plot/.style={
		axis on top,
		clip marker paths=true,
		scale only axis,
		height=\matrixrows/\matrixcols*\pgfkeysvalueof{/pgfplots/width},
		enlarge x limits={rel=0.5/\matrixcols},
		enlarge y limits={rel=0.5/\matrixrows},
		scatter/use mapped color={draw=mapped color, fill=mapped color},
		scatter,
		point meta=explicit,
		mark=square*,
		cycle list={
			mark size=0.5*\pgfkeysvalueof{/pgfplots/width}/\matrixcols
		}
	},
	matrix rows/.store in=\matrixrows,
	matrix rows=8,
	matrix cols/.store in=\matrixcols,
	matrix cols=10
}

\tikzset
{
	myTrapezium/.pic =
	{
		\draw [fill=blue2!30, thick] (0,0) -- (0,\b) -- (\a,\c) -- (\a,-\c) -- (0,-\b) -- cycle ;
		\coordinate (-center) at (\a/2,0);
		\coordinate (-out) at (\a,0);
	},
}

\tikzset
{
	smallTrapezium/.pic =
	{
		\draw [fill=blue2!30, thick] (0,0) -- (0,\bs) -- (\as,\cs) -- (\as,-\cs) -- (0,-\bs) -- cycle ;
		\coordinate (-center) at (\as/2,0);
		\coordinate (-out) at (\as,0);
	},
}

\tikzset{
  pics/codebook/.style={
    code={
      \coordinate (codebook) at (0,0);
      \node[draw,rounded corners,minimum width=0.41*\rectSize,minimum height=0.4*\rectSize,thick] (crect) at (codebook) {};
      
      \node at ($(codebook)+(0,0.53)$) {\scriptsize Codebook $\mathcal{Z}$};

      \node[anchor=east,draw,inner sep=0.8pt,fill=white,minimum width=0.35cm,minimum height=0.18cm] (cbl0) at ($(-0.08*\rectSize,0.28)$) {\tiny 1};
      \node[anchor=west,draw,inner sep=0.8pt,fill=blue0,minimum width=0.9cm,minimum height=0.18cm] (cbr0) at (cbl0.east) {};

      \node[anchor=north,draw,inner sep=0.8pt,fill=white,minimum width=0.35cm,minimum height=0.18cm] (cbl1) at (cbl0.south) {\tiny 2};
      \node[anchor=west,draw,inner sep=0.8pt,fill=green0,minimum width=0.9cm,minimum height=0.18cm] (cbr1) at (cbl1.east) {};

      \node[anchor=north,draw,inner sep=0.8pt,fill=white,minimum width=0.35cm,minimum height=0.18cm] (cbl2) at (cbl1.south) {\tiny 3};
      \node[anchor=west,draw,inner sep=0.8pt,fill=orange0,minimum width=0.9cm,minimum height=0.18cm] (cbr2) at (cbl2.east) {};

      \node[anchor=north,draw=white,inner sep=0.8pt,fill=white,minimum width=0.35cm,minimum height=0.18cm] (cbl3) at (cbl2.south) {};
      \node[anchor=west,draw=white,inner sep=0.8pt,fill=white,minimum width=0.9cm,minimum height=0.18cm] (cbr3) at (cbl3.east) {$\dots$};

      \node[anchor=north,draw,inner sep=0.8pt,fill=white,minimum width=0.35cm,minimum height=0.18cm] (cbl4) at (cbl3.south) {\tiny Q};
      \node[anchor=west,draw,inner sep=0.8pt,fill=blue0!60,minimum width=0.9cm,minimum height=0.18cm] (cbr4) at (cbl4.east) {};
    }
  }
}

\newcommand\ssW{1cm}     
\newcommand\ssH{0.8cm}   
\tikzset{
  pics/shadowstackfixed/.style={
    code={
      \node[anchor=center, draw, thick, fill=white,
            minimum width=\ssW, minimum height=\ssH] (-layer1) at ( 2pt,  4pt) {};
      \node[anchor=center, draw, thick, fill=white,
            minimum width=\ssW, minimum height=\ssH] (-layer2) at ( 1pt,  2pt) {};
      \node[anchor=center, draw, thick, fill=white,
            minimum width=\ssW, minimum height=\ssH] (-layer3) at ( 0pt,  0pt) {};
      \node[anchor=center, draw, thick, fill=white,
            minimum width=\ssW, minimum height=\ssH] (-top) at (-1pt, -2pt) {};
    }
  }
}
\def\a{0.7}  
\def\b{.3} 
\def\c{0.45}  

\def\as{0.7}  
\def\bs{.2} 
\def\cs{0.35}  

\begin{tikzpicture}
\def \radS {2.5mm}
\coordinate (x0) at (0,0);
\coordinate (g0) at (-0.7, +0.1);;
\def \rectSize {3.5cm}
\def \descriptiveSize {3cm}
\coordinate (c0) at ($(x0) + (3.35, 0)$);
\coordinate (c1) at ($(c0) + (0.5*\rectSize, 0)$);

\coordinate (nn0) at ($(c0)+ (0.95*\rectSize,0)$);
\coordinate (nnOUT) at ($(nn0) + (3.5,0)$);

\node
(obs) at ($(x0)+ (4.5cm,-0.95)$) {};

\node[draw=white](Yzero) at ($(x0)+(-0.25, 0)$) {$\mathbf{Y}$};

\node [anchor= west, align =center, draw=black,  fill= gray, fill opacity= 0.1, text opacity=1,  shape=rectangle, rounded corners,text width = 0.9cm, minimum width=0.32*\rectSize, minimum height=0.2*\rectSize] (invvmm) at ($(Yzero)+(1.0,0)$) {\small $\frac{d}{dt}$};

\draw [ ->] (Yzero) -- (invvmm) node [pos=.5, above] (TextNode1) {};
\draw[->] (invvmm) -- (c0) node[pos=0.4, fill=white, inner sep = 1pt, ] (ix0) {\small $\bm{x}_0$};

\node [anchor= west, draw, fill=red1, fill opacity= 0.1, text opacity=1,  shape=rectangle,rounded corners, minimum width=0.9*\rectSize, minimum height=0.16*\rectSize] (fd) at (c0) {Forward Diffusion};
\node [anchor= west, draw,  fill= red1, fill opacity= 0.1,text opacity=1, shape=rectangle, rounded corners, minimum width=0.9*\rectSize, minimum height=0.16*\rectSize] (rd) at ($(c0)+(0,-0.8)$) {Reverse Diffusion};
\draw [->, shorten >=6pt] (rd.west) -- ($(ix0) + (0, -0.8)$)  node [pos=1, fill=white, inner sep = 1pt] (TextNode1) (hx0) {\small $\bm{\hat{x}}_{0}$};
\draw[<->, dotted, thick, red1] (ix0) -- (hx0) node[] {};
\begin{scope}[on background layer]
\node[
    draw= red1,
    fill = red1!20,
    dotted,
    thick,
    rounded corners=0.2cm,
    fit=(fd)(rd),     
    inner sep=6pt,
] (diffusion_group) {}; 
\end{scope}

\coordinate (mid_fd_rd_right) at ($ (fd)!0.5!(rd) + (3, -0.3) $);
\node (xt_label) at (mid_fd_rd_right) {\small$\bm{x}_T$};
\begin{axis}[
name=xtaxis,
anchor={south},
at={($(mid_fd_rd_right)$)},
yshift = 0.3cm,
axis lines=center,
axis line style=thin,
enlargelimits=false, 
ticks=none,
width=2.5cm,
height=2.1cm,
xmin=-4, xmax=4,
ymin=-0.02, ymax=0.6,
clip=false
]
\addplot[black!50,thick,name path=B,domain={-3}:{3},samples=100,smooth,] {gauss(x,0,1)};
\end{axis}

\begin{scope}[on background layer]
\node[
    draw=black,
    rounded corners,
    fill=white,
    fit=(xtaxis)(xt_label),   
    inner sep=6pt
] (xt_group) {};
\end{scope}
\draw [<->] (diffusion_group.east) -- (xt_group.west);



\node [anchor= west, draw, text width = 0.8*\rectSize, align=center, fill= green1!10, fill opacity = 0.1, text opacity=1,  shape=rectangle, rounded corners, minimum width=0.9*\rectSize, minimum height=0.16*\rectSize] (bwd) at ($(c0)+(0,-2.25)$) {Generative Diffusion};
\begin{scope}[on background layer]
\node[
    draw=green1,
    fill = green1!10,
    dotted,
    thick,
    rounded corners=0.2cm,
    fit=(bwd),     
    inner sep=6pt,
] (inference_group) {};
\end{scope}

\draw [->] (xt_group.south) |-($(inference_group.east) + (0, 0.2)$);

\node [draw=black,  fill= gray, fill opacity= 0.1, text opacity=1, shape=rectangle, rounded corners, minimum width=0.25*\rectSize, minimum height=0.2*\rectSize] (vmm) at ($(invvmm |- bwd)$) {VMM};
\draw [->] (bwd.west) -- (vmm.east)  node [pos=.6, fill=white] (TextNode1) {\small $\bm{\hat{x}}_{0}$};

\draw[->] (vmm.west) -- ($(Yzero.east |- vmm)$) node[left] (xTb) {$\mathbf{\hat{Y}}$};
\draw[<->, thick, dotted, thick, green2] (Yzero) -- (xTb) node[] {};

\begin{scope}[on background layer]
\node[
    draw=gray,
    thick,
    fill = white,
    fill opacity = 0.1,
    rounded corners,
    fit=(diffusion_group)(xt_group)(inference_group),   
    inner sep=6 pt,
    minimum height=4cm,
    label={[above, yshift=-3pt]north: \small DDPM + DDIM sampling} 
] (fef) {};
\end{scope}

\node[] (cond) at ($(fef.east) + (0.25*\columnwidth, +0.625)$) {};

\pic (cb) at (cond) {codebook};

\draw[very thick, <-, blue]
  ($(fef.east |- cond)$)
  -- ($(cond)+(-0.7,0)$)
  node[pos=0.4, fill=white] (inputc) {\color{black} $c$};

\coordinate (coorzhat) at ($(cond.east) + (3,0.75)$);
\coordinate (coorzh) at ($(cond.east) + (3,-0.75)$);

\node (zq) [anchor= center, 
shape=rectangle,
draw, thick,
fill=blue2!60,
minimum width = 0.6 cm, 
minimum height=2*\b cm,
] at (coorzh) {$z_q$};

\def \dh {0.10};
\draw[] ($(zq.north west)+(0.02, -0.005)$) -- ++ (\dh, \dh) --++ (0.6cm, 0) -- ++ (-\dh, -\dh);
\draw[] ($(zq.south east)+(-0.005, 0.02)$) -- ++ (\dh,\dh) --++ (0, 2*\b cm);

\node (zhat)[anchor= center, 
shape=rectangle,
draw, thick,
fill=blue2!60,
minimum width = 0.6  cm, 
minimum height=2*\b cm,
] at (coorzhat) {$\hat{z}$};
\draw[] ($(zhat.north west)+(0.02, -0.005)$) -- ++ (\dh, \dh) --++ (0.6cm, 0) -- ++ (-\dh, -\dh);
\draw[] ($(zhat.south east)+(-0.005, 0.02)$) -- ++ (\dh,\dh) --++ (0, 2*\b cm);

\coordinate (mid_step) at ($(zhat.east)-(1.25cm,0)$);
\coordinate (mid_up)   at ($(mid_step |- zq)$);
\draw[->] (zhat.west) -- (mid_step) -- (mid_up) -- (zq.west);
\node[
  font=\scriptsize,
  align=center,
  text width=1.2cm,
  fill = white,
  inner sep = 0.01pt,
] (cc) at ($(mid_step)!0.5!(mid_up)$)
{$\arg\min$ \\ $||\bm{\hat{z}} - \bm{z}_k||$};

\draw[<->,  thick, black]
  (cc.west) -- ($(cond.east) + (0.6,0)$)
  node[pos=0.5] (cccond_center) {};
\pic (right) at ($(coorzhat)+(0.35, 0)$) {myTrapezium};
\node (E) at (right-center) {$E$} ;

\pic (left) at ($(coorzh)+(0.35, 0)$) {myTrapezium} ;
\node (D) at (left-center) {$D$};

\coordinate (inputxi) at ($(right-center) + (1.5,0)$);
\coordinate (outputxi) at ($(left-center) + (1.5,0)$);
\pic (xi) at (inputxi) {shadowstackfixed};
\node[] at (inputxi) {${\bm{\xi}}$};
\draw [<-, ] ($(coorzhat)+(0.35, 0)+(\a, 0)$) -- ($(inputxi) -(0.55, 0)$) ;
\pic (hxi) at (outputxi) {shadowstackfixed};
\node[] at (outputxi) {$\hat{\bm{\xi}}$};
\draw [->] ($(coorzh)+(0.35, 0)+(\a, 0)$) -- ($(outputxi) -(0.55, 0)$);

\draw[<->, dotted, thick, red1, shorten >=14pt, shorten <=16pt] (outputxi) -- (inputxi) node[] {};

\coordinate (W) at ($(inference_group.east)+(0,-0.2)$);
\node [draw=black,  align=center, fill= gray, fill opacity= 0.1, text opacity=1, shape=rectangle, rounded corners, text width=0.35*\rectSize, minimum height=0.1*\rectSize] (uq) at ($(cond |- W)$) {UQ};

\node[] (sigma) at ($(inputc |- uq)$) {$\delta$};
\draw[->, very thick, blue]  (uq.west) -- (sigma);
\draw[->, very thick, blue]  (sigma) -- (W);

\draw [<-, dashdotted, thick] 
    (uq.north) -- ($(cond) + (0, -0.7)$)
    node[pos= 0.6, fill=white, inner sep =1.5pt, ] {\small $\bm{z}_q$};

\draw [->, dashdotted,thick] (cccond_center) |- (uq.east);

\begin{scope}[on background layer]
\node[
  draw=gray,
  thick,
  rounded corners,
  minimum width=5.5cm,
  minimum height=2.75cm,
  anchor=center,
  label={[above, yshift=-3pt]north: \small CVQ-VAE} 
] (vqcae) at ($(cond)+(1.75,0)$) {};   
\end{scope}

\coordinate (cl) at ($(coorzh) + (0.35,-1.35)$);
\pic (fcl) at (cl.west) {smallTrapezium};

\draw [->, ] (zq.south) |- (cl);

\node at (fcl-center) {$f_{cl}$} ;
\coordinate (pi) at ($(inputxi |- cl)$);
\draw [->, ] ($(cl) +(\as,0)$) -- ($(pi)+(-0.2,0)$) ;
\node at (pi) {$\bm{p}$} ;

\node [anchor= west, rotate=90, draw, inner sep=0.8pt, fill= white, shape=rectangle,  minimum width=0.9cm, minimum height=0.25cm] (sftmx) at ($(cl)+ (0.9, -0.45)$) {\scriptsize  softmax};

\def\bottom{3.5cm} 
\coordinate (L) at ($(Yzero) + (0.5,-\bottom)$);
\coordinate (R) at ($(Yzero) + (9.5cm,-\bottom)$);
\draw[very thick, gray!80] (L) -- (R)
  node[midway, below] {\small Vehicle Motion Diffusion};
\draw[very thick, gray!80] (L) --++  (0, 10pt);    
\draw[very thick, gray!80] (R) --++ (0, 10pt);     

\coordinate (LL) at ($(R) + (0.25,0)$);
\coordinate (RR) at ($(LL) + (6.25cm,0)$);
\draw[very thick, gray!80] (LL) -- (RR)
  node[midway, below] {\small Context Conditioning};
\draw[very thick, gray!80] (LL) --++  (0, 10pt);    
\draw[very thick, gray!80] (RR) --++ (0, 10pt);     

\end{tikzpicture}

%% file: tables/Ablation_study.tex
\begin{tabular}{l c ccccccc}
\toprule
\multirow{2}{*}{Mode} 
& \multirow{2}{*}{Metric} 
& \makecell{VQ-VAE\cite{Oord2017, Neumeier2024}} 
& \multicolumn{5}{c}{CVQ-VAE \cite{Zheng2023}} \\
\cmidrule(lr){3-3}\cmidrule(lr){4-8}
& 
& $Q=60$ & $Q=30$ & $Q=60$ & $Q=90$ & $Q=128$ & $Q=256$ \\
\midrule
\midrule
Mean                 & ADE [\SI{}{\meter}] $\downarrow$                     & 1.44 & 1.42 & 1.40 & 1.41 & 1.41   & \textbf{1.37} \\
Mean                  & FDE [\SI{}{\meter}]@\SI{5}{\second}  $\downarrow$   & 3.99 & 3.94 & 3.90 & 3.92 & 3.92  & \textbf{3.84} \\ \hline
Most Likely $p(c)$   & ADE [\SI{}{\meter}] $\downarrow$                     & 1.46   & 1.45  & 1.44   & 1.45   & 1.45  & \textbf{1.43}   \\
Most Likely  $p(c)$ & FDE  [\SI{}{\meter}]@\SI{5}{\second}  $\downarrow$    & 4.14  & 4.04  & 4.02   &  4.09   & 4.03  & \textbf{4.00}   \\ \hline
Best-of-$K$   & MinADE$_k$ [\SI{}{\meter}] $\downarrow$                     & 1.37 & 1.35 & 1.34 & 1.35 & 1.34   & \textbf{1.33} \\
Best-of-$K$   & MinFDE$_k$ [\SI{}{\meter}]@\SI{5}{\second}  $\downarrow$    & 3.79 & 3.75 & 3.72 & 3.80 & 3.72   & \textbf{3.68} \\
\bottomrule
\end{tabular}

%% file: tikz-figures/dist_approx.tex
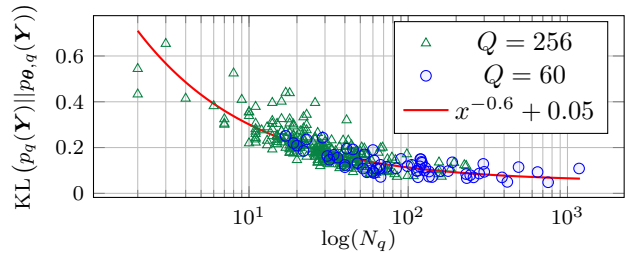
\begin{figure}[t]
\begin{tikzpicture}
\begin{axis}[
    width=\columnwidth,
    height=4.0cm,
    xlabel={ \footnotesize $\log(N_q)$},
    ylabel={ \footnotesize $\mathrm{KL}\left(p_q(\bm{Y})||p_{\bm{\theta}, q}({\bm{Y}})\right)$}, 
    xmode=log,
    ticklabel style={font=\footnotesize},   
    xlabel style={yshift=5pt},
    legend style={at={(0.97,0.97)}, anchor=north east},
    grid=both
]

\addplot[
    only marks,
    mark=triangle,
    draw=green1,
    fill=green1,
    fill opacity=0.05
] table[
    x=data_length,
    y=JS,
    col sep=comma
] {data/dist_results_256.csv}; \label{q256}
\addlegendentry{$Q=256$}

\addplot[
    only marks,
    mark=*,
    draw=blue,
    fill=blue,
    fill opacity=0.05
] table[
    x=data_length,
    y=JS,
    col sep=comma
] {data/dist_results_60.csv}; \label{q60}
\addlegendentry{$Q=60$}

\addplot[
    thick,
    red,
    domain=2:1200,
    samples=200,
]
{1 * x^(-0.6) + 0.05}; \label{approx}

\addlegendentry{$x^{-0.6} + 0.05$};
\end{axis}
\end{tikzpicture}
\vspace{-5pt}
\caption[Caption in ToC]{Average KL divergence as a function of the number of samples for 
$Q=60$ (\ref{q60}) and $Q=256$ (\ref{q256}). Larger $Q$ reduces the per-cluster sample count $N_q$, yielding poorer distribution estimates and higher KL divergence. The empirical relationship can be approximated by $x^{-0.6} + 0.05$ (cf. \ref{approx}).}
\label{fig:kld}
\end{figure}

%% file: tables/Performance_evaluation.tex
\begin{tabular}{c|c|c}
		\multirow{2}{*}{Architecture}&
		\multicolumn{2}{c}{highD} \\
		&ADE [\SI{}{\meter}] $\downarrow$ 
		& FDE [\SI{}{\meter}]@\SI{5}{\second}  $\downarrow$\\
		\hline
		\hline
		GFTNNv2 \cite{Neumeier.ITSC2023}&  \textbf{0.72} & \textbf{1.80}  \\	 
		HSTA \cite{HSTA.2021}        	& 2.18 & 4.56 	\\	
		CS-LSTM	 \cite{Deo.2018}	 	& 2.88 & 5.71\\	 
		MHA-LSTM(+f) \cite{Messaoud2021} & 2.58 & 5.44 	\\	 
		Two-channel \cite{Mo2021}  & 2.97 & 6.30 	\\	 
		cVMD \cite{Neumeier2024}   & 1.79 & 3.76 \\  \hline 
        \textbf{cVMDx$^{\textrm{(mean)}}$}   & 1.37 & 3.84 \\  
        \textbf{cVMDx$^{(\bm{p(c)})}$}   & 1.43 & 4.00 \\ \hline 
\end{tabular}%